\documentclass[10pt,twocolumn,letterpaper]{article}

\usepackage{iccv}
\usepackage{times}
\usepackage{epsfig}
\usepackage{graphicx}
\usepackage{amsmath}
\usepackage{amssymb}

\usepackage{array}
\usepackage{float}

\usepackage{graphicx} 
\usepackage{float} 
\usepackage{subfigure} 

\usepackage{authblk}

\usepackage{makecell}

\usepackage{enumitem}
\setenumerate[1]{itemsep=0pt,partopsep=0pt,parsep=\parskip,topsep=5pt}
\setitemize[1]{itemsep=0pt,partopsep=0pt,parsep=\parskip,topsep=5pt}
\setdescription{itemsep=0pt,partopsep=0pt,parsep=\parskip,topsep=5pt}

\usepackage[colorlinks,
            linkcolor=blue,
            anchorcolor=blue,
            citecolor=blue]{hyperref}

\usepackage{hyperref}
            
\iccvfinalcopy 

\def\iccvPaperID{****} 

\ificcvfinal\fi

\begin{document}

\title{Boosting Generalization with Adaptive Style Techniques for Fingerprint Liveness Detection}

\author[]{Kexin Zhu}
\author[]{Bo Lin}
\author[]{Yang Qiu}
\author[]{Adam Yule}
\author[]{Yao Tang}
\author[]{Jiajun Liang}

\affil[]{JIIOV Technology\thanks{Jiiov Technology \url{https://jiiov.com}}}
\affil[]{\tt\normalsize{\{kexin.zhu, bo.lin, yang.qiu, mochen.yu, yao.tang, jiajun.liang\}@jiiov.com}}

\maketitle
\ificcvfinal\thispagestyle{empty}\fi

\begin{abstract}

  We introduce a high-performance fingerprint liveness feature extraction technique that secured \textbf{first place in LivDet 2023 Fingerprint Representation Challenge}. Additionally, we developed a practical fingerprint recognition system with 94.68\% accuracy, earning second place in LivDet 2023 Liveness Detection in Action. By investigating various methods, particularly style transfer, we demonstrate improvements in accuracy and generalization when faced with limited training data. As a result, our approach achieved state-of-the-art performance in LivDet 2023 Challenges.
\end{abstract}

\section{Introduction}

Fingerprint-based authentication systems have become increasingly important in a variety of fields such as electronic payment and access control, owing to their security and efficiency. Despite their advantages, these systems remain vulnerable to attacks involving counterfeit fingerprints created from materials like silica, gelatin, and latex. Such fake fingerprints can deceive the system by mimicking genuine ones. Consequently, it is crucial to develop liveness detection capabilities for bolstering system robustness and security.

A typical liveness detection algorithm encompasses image processing, feature extraction and classification, with the complete system incorporating finger matching and liveness detection algorithms. Fingerprint representation is critical in detecting attacks, and effective representation extraction methods should consider discriminability, compactness and speed as key factors. LivDet 2023\cite{micheletto2023livdet2023} Challenge 2 emphasizes fingerprint representation tasks.

Moreover, Fingerprint Liveness Detection systems are intended to function as part of a broader recognition system, where recognition algorithms aid in identifying low-quality attacks. Therefore, evaluating system performance takes into account integrated match scores, including template matching and liveness scores. LivDet 2023 Challenge 1 pertains to this task.

In this paper, we tackle both challenges by investigating a fingerprint liveness representation extraction algorithm and developing a detection system with enhanced generalization. Our three main contributions include:
\begin{itemize}
	\item We explore a feature extraction method with advancing generalization performance and low latency simultaneously, which better handles the bias-variance trade-off.
	\item We propose a practical Fingerprint Recognition System, which is an end-to-end solution to differentiate users from non-users and genuine from attacks.
	\item Based on the above method, we won the first place and second place in LivDet 2023 Challenge 2 and Challenge 1 respectively.
\end{itemize}

\section{Methods}


In this section, we initially examine strategies to address the bias-variance trade-off\cite{belkin2019reconciling}\cite{neal2018modern} in order to achieve reduced test error. This involves five key design choices for our method: increased model capacity, strong augmentation, style transfer, mutual learning and model ensemble. Subsequently, we present an efficient feature extraction approach aimed at minimizing time consumption. Lastly, we illustrate the construction of a liveness detection system, consisting of a liveness monitor and recognition matcher, which can effectively improve the accuracy of detecting fake fingerprints.

\begin{figure*}[ht] 
\centering 
\includegraphics[scale=0.47]{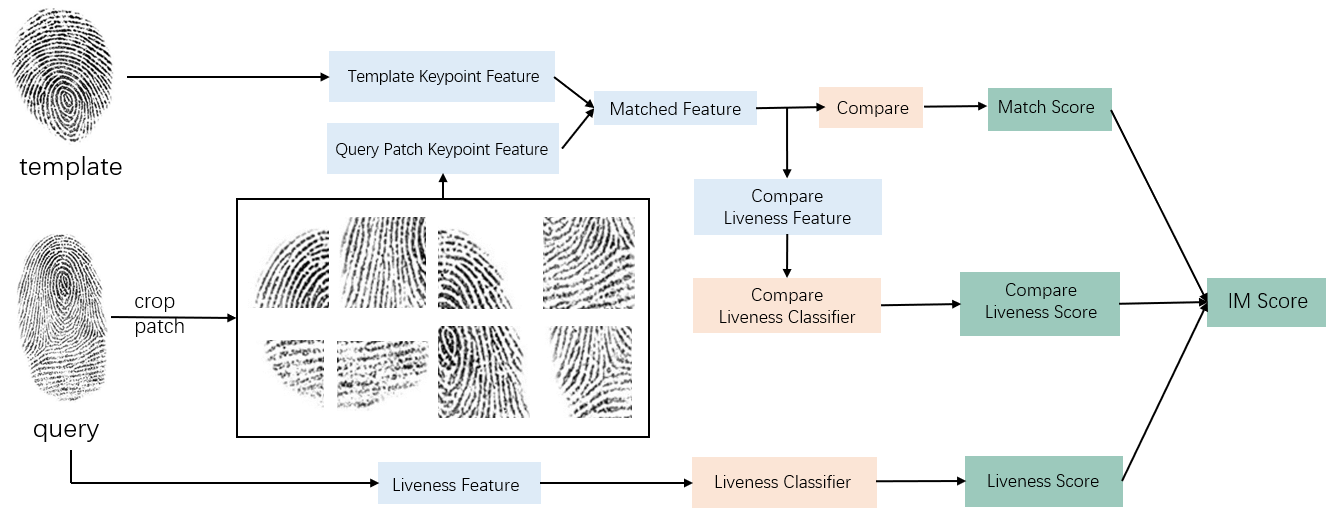} 
\caption{Fingerprint recognition system. The system comprises three components. Firstly, the comparison module takes the template and queries the patch keypoint feature, subsequently comparing them to produce a match score that evaluates the degree of similarity between the two fingers. Secondly, the compare liveness classifier outputs liveness score based on matched features. Lastly, liveness classifier extract query feature to output the normal liveness score. The ultimate output IM score is integrated by match score, compare liveness score and normal liveness score.} 
\label{Fig_system} 
\end{figure*}

\subsection{How to handle with Bias-variance trade-off}

\noindent\textbf{Increase model capacity to reduce bias.} The bias results from the inability of the model to describe the true underlying function. This suggests that we can reduce bias error by making the model more flexible\cite{prince2023understanding}. Large models successively refresh the state\text{-}of\text{-}art among various tasks in both vision and language fields. Thus we study the effect of model capacity on the LivDet dataset. We conduct experiments with four different-sized models, namely MobileNetV3 Small, MobileNetV3 Large, SE\text{-}ResNeXt50, SE\text{-}ResNeXt101\cite{koonce2021mobilenetv3}\cite{xie2017aggregated}, ranging from small to large. As the model size grows, the accuracy improves, while the boost speed decreases. In consideration of minor improvement from SE\text{-}ResNeXt50 to SE\text{-}ResNeXt101, we select SE\text{-}ResNeXt50 as backbone for better accuracy\text{-}speed trade-off.

\noindent\textbf{Stronger augmentation help generalize better.} Due to the limited amount of training data, strong augmentation can significantly enhance data diversity, thus reducing the discrepancy between the accuracy of the training and validation sets. Geometric augmentations, such as horizontal and vertical flips, translation, cropping, affine transformations and rotation are employed. Color augmentations encompass adjustments to lightness and contrast jitter\cite{imgaug}. Fmix\cite{harris2020fmix}, a mixed-sample data augmentation method that utilizes random binary masks derived from low-frequency images, plays a crucial role. All the augmentations are randomly permuted and applied with a probability of 0.5\cite{jung2019imgaug}\cite{papakipos2022augly}.

\noindent\textbf{Style transfer make model adapt to distinct source of data.} The factors that significantly influence fingerprint style include the user, material and scanner. In the realm of domain adaptation\cite{csurka2017domain}\cite{farahani2021brief}, it is well-established that image mean and variance can effectively characterize an image's style. Given a content image and a style reference image, style swapping allows the content image retain its inherent content while adopting the style of the reference image. \cite{tang2021crossnorm}.  Analogously, we swap styles between two samples during the training process to generate greater diversity. It is worth to emphasize that the swapping occurs only between data with same label, as we think the style of genuine and attack differs. If cross-label swapping were to occur, it would undermine discriminative characteristics. Consequently, the ability to generalize on unseen users, materials and scanners is enhanced thanks to style transfer technique.

\noindent\textbf{Mutual learning alleviate over-confident.} Models trained exclusively with softmax and cross-entropy loss often exhibit overconfidence, which is harmful to generalization. In addition to cross entropy loss, we simultaneously train two identical networks by minimizing the Kullback-Leibler (KL) divergence between their predictions. Owing to the influence of cross-entropy loss and random initialization, the two models possess distinct prediction distributions, apart from the target class. Consequently, mutual learning enables the model to focus not only on the target class prediction but also on the probability distribution. This approach facilitates convergence to a flat and smooth minimum instead of a narrow and sharp one, thereby providing a regularization effect\cite{zhang2018deep}.

\begin{figure}[htbp] 
\centering 
\includegraphics[scale=0.5]{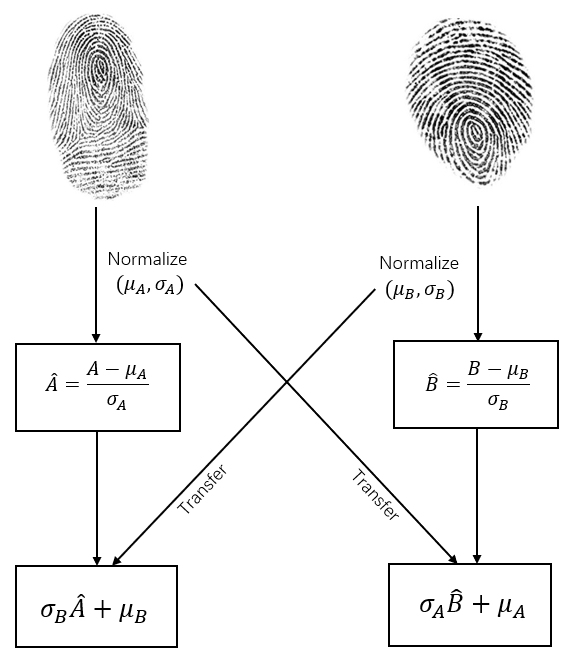} 
\caption{Style transfer. During training process, the mean and standard deviation statistics are swapped between two samples in order to increase the diversity of the training data.} 
\label{Fig_style_transfer} 
\end{figure}

\noindent\textbf{Over\text{-}parameterized models suffer from overfit but ensemble continue to help.} Increasing the model capacity does not evidently improve performance, as the over\text{-}parameterized model overwhelms the dataset size and tends to overfit. Ensembling multiple model predictions is one way to increase computational cost and boost the performance\cite{dietterich2000ensemble}. Final outputs refer to multi-view predictions, which can stabilize the prediction and reduce variance. We employ an ensemble of three models, comprising a style transfer model and two mutual learning models, all of which use SE\text{-}ResNeXt50 as their backbone.

\subsection{Efficient Representation Extraction}

\noindent\textbf{Knowledge Distillation.} We train SE\text{-}ResNeXt101 models to transfer the knowledge to SE\text{-}ResNeXt50 models for further improvement. We use temperature=5 to generate the soft prediction. The student network is trained by cross entropy loss and Kullback-Leibler (KL) divergence loss between soft predictions of teacher and student models\cite{gou2021knowledge}. Consequently, our proposed model demonstrates remarkable generalization capabilities while maintaining low inference latency.

\subsection{Fingerprint Recognition System}
An end-to-end fingerprint recognition system is capable of not only detecting attacks but also matching fingerprints, making it more practical for real\text{-}world applications. Our proposed system begins by preprocessing the query image, followed by dividing it into patches. Keypoint features are then extracted from each patch and the template\cite{shreyas2017fingerprint}\cite{shuai2008fingerprint}. The keypoint features of each patch are compared to those of the template, resulting in a patch match score. Simultaneously, liveness features are extracted based on the matched features and passed through a liveness classifier to obtain a liveness comparison score. The final match score and liveness comparison score are both ensembled by all patches. We also encode the query liveness feature to obtain a normal liveness score. Ultimately, the final score, referred to as the IM score, is an integrated value obtained by combining the match score, liveness comparison score and normal liveness score. The design of the system is shown as Figure~\ref{Fig_system}.

\section{Results}

\subsection{Evaluation Metric}
In both challenges, the performance of the Fingerprint Presentation Attack Detection (PAD) will be evaluated using the standard PAD ISO metrics\cite{iso2022iso}\cite{iso2023iso}:
\begin{itemize}
	\item PAD Accuracy: percentages of fingerprint images correctly classified by the PAD.
	\item BPCER (Bona fide Presentation Classification Error Rate): Rate of misclassified bona fide images.
	\item APCER (Attack Presentation Classification Error Rate): Rate of misclassified fake images.
\end{itemize}

In Challenge 1, to evaluate the performance of the integrated system, the following metrics are employed\cite{micheletto2023livdet2023}:
\begin{itemize}
	\item FNMR (False Non-Match Rate): Rate of mated comparisons that result in rejection.
	\item IAPAR (Impostor Attack Presentation Accept Rate): rate of presentation attacks that result in acceptance.
	\item Integrated Matching (IM) Accuracy: percentages of samples correctly classified by the integrated system.
\end{itemize}

\begin{figure}[htbp] 
\centering 
\includegraphics[scale=0.5]{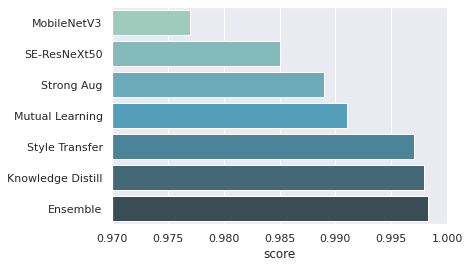} 
\caption{An overview of all methods AUC score on validation set.} 
\label{Fig_trick} 
\end{figure}

\begin{figure*}[ht] 
\centering 
\includegraphics[scale=0.5]{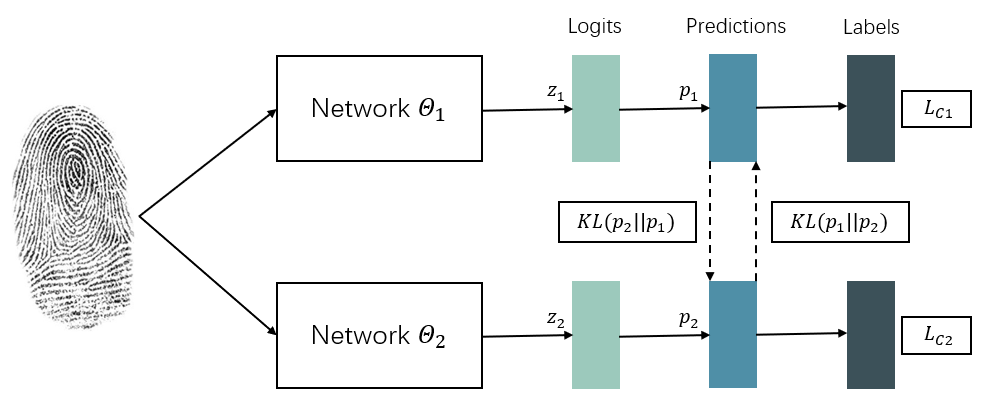} 
\caption{Mutual learning. The two networks share the identical structure and are randomly initialized. Both are optimized by cross-entropy loss and Kullback-Leibler (KL) divergence between predictions, which can alleviate overconfidence and improve generalization capabilities.} 
\label{Fig_mutual_learning} 
\end{figure*}

\subsection{Challenge2 Result}
In this section, we will first describe our experiments mentioned above and then present the results of Challenge 2.

The dataset is divided into training set and validation set, accounting for 2/3 and 1/3 of the dataset respectively. Both subsets are randomly sampled. We adopt MobileNetV3 and simple augmentation as our baseline. Larger model SE-ResNeXt50 brings about 0.8\% AUC increase from 97.7\% to 98.5\% on validation set. Strong augmentation is beneficial by further improving AUC to 98.9\%. Subsequently, we conduct mutual learning experiment and observe a moderate improvement of 0.2\%. Style transfer demonstrates its effectiveness by boosting AUC with a remarkable increase from 98.9\% to 99.7\%, while knowledge distillation exhibits a marginal increment of 0.1\%. At inference time, we ensemble the style transfer model and two mutual learning models, resulting in a mild growth to 99.83\%. The overall results are illustrated in Figure~\ref{Fig_trick}.

The public results of Challenge 2 is demonstrated in Table~\ref{table_ch2_result}. Our proposed algorithms are denoted as \textit{jiiov} and \textit{jiiov\underline{~}all}. In order to determine the final score, we comprehensively consider factors including PAD accuracy, inference time and feature size. As depicted in Table~\ref{table_ch2_result}, our method attains the optimal trade-off among these factors, rendering it more appropriate for real-world applications and consequently winning the challenge. It is necessary to mention that the \textit{jiiov\underline{~}all} algorithm's model is trained with both LivDet 2023 and LivDet 2021 datasets. However, it exhibits reduced accuracy on test set. This phenomenon primarily suggests the presence of a domain gap between the two data sources due to unknown attack materials and scanners. The aforementioned gap adversely affects the model's performance on the LivDet 2023 benchmark.

\subsection{Challenge1 Result}
As is demonstrated in Table~\ref{table_ch1_result}, our algorithm achieves 94.68\% IM accuracy, 1.7\% falling behind the champion team, ranking the second place. The primary factor contributing to this disparity is the relatively inferior liveness performance of our model, which exhibits a PAD accuracy that is 7.4\% lower than that of the leading team. Although we consider the efficiency of our model, it appears not to be a considered metric in Challenge 1. The liveness model is hindered by an inadequate representation capability.

\section{Conclusion}
In this report, we explore an effective and efficient fingerprint liveness feature extraction method that advances generalization performance and better handles the bias-variance trade-off. Meanwhile we achieve the optimal performance in terms of latency, feature size and accuracy trade-off and thus won the first place in Challenge 2. We propose a practical fingerprint recognition system, which is an end-to-end solution for accepting real users while rejecting non-users or counterfeit fingers. The system accuracy achieved 94.68\%, earning second place in Challenge 1.

Despite exploring methods to improve generalization and consequently achieving the highest score, we discovered a significant accuracy drop from validation set to test set due to unknown attack materials, scanners and users. The development of a generalized liveness classifier remains exceedingly challenging and there is still a long way to go.

\section{Acknowledgements}
We would like to convey 
 heartfelt gratitude to Jiiov Technology for their invaluable support throughout this research project. Special thanks go to the mobile hardware development team, software development team, algorithm performance testing team and high-performance computing team for their tireless efforts and dedication, which have contributed significantly to the success of this study.

\begin{table}[htbp]   
\centering
\caption{Challenge 2: overall results.}  
\label{table_ch2_result} 
\scalebox{0.9}{
\begin{tabular}{|c|c|c|c|c|}   
\hline   \textbf{Algo} & \thead{\textbf{Overall} \\ \textbf{Time[ms]}} &  \thead{\textbf{Feat} \\ \textbf{size}} & \textbf{Acc[\%]} & \textbf{Score} \\   
\hline   Contr1 & 1302.97 & 800 & 87.65 & 0.57  \\ 
\hline   Contr2 & 4511.78 & 800 & 79.03 & 0.00  \\  
\hline   unina2 & 93.80 & 32 & 79.80 & 0.69 \\
\hline   unina3 & 94.10 & 32 & 80.70 & 0.73 \\   
\hline   \pmb{jiiov} & 46.89 & 192 & 84.29 & \pmb{0.80} \\
\hline   jiiov\underline{~}all & 47.42 & 192 & 80.55 & 0.66 \\   
\hline   
\end{tabular}}
\end{table}

\begin{table}[htbp]   
\centering
\caption{Challenge 1: overall results.}
\label{table_ch1_result} 
\scalebox{0.8}{
\begin{tabular}{|c|c|c|}   
\hline   \textbf{Algorithm} & \thead{\textbf{Overall PAD} \\\textbf{Accuracy [\%]}} & \thead{\textbf{Overall IM} \\\textbf{Accuracy [\%]}} \\   
\hline   Contr1 & 92.47 & 53.69   \\ 
\hline   Contr2 & 87.42 & 42.09   \\  
\hline   CIS\underline{~}F & 88.75 & 91.55  \\
\hline   CIS\underline{~}W & 96.22 & 95.99  \\   
\hline   \pmb{CIS\underline{~}Wens} & \pmb{97.54} & \pmb{96.35} \\
\hline   CIS\underline{~}F\underline{~}v2 & 89.22 & 91.2  \\   
\hline   S\underline{~}cls & 93.76 & 94.11  \\
\hline   S\underline{~}knn & 94.05 & 94.11  \\
\hline   HNU\underline{~}AIM & 75.8 & 86.29  \\
\hline   unina1 & 86.24 & 93.14   \\ 
\hline   unina2 & 87.42 & 92.86   \\
\hline   unina3 & 88.66 & 93.47   \\ 
\hline   unina4 & 87.23 & 84.87   \\
\hline   unina5 & 86.65 & 93.07   \\
\hline   unina6 & 88.3 & 93.06   \\
\hline   \pmb{jiiov} & \pmb{90.12} & \pmb{94.68}   \\
\hline   jiiov\underline{~}all & 87.11 & 94.33   \\
\hline   
\end{tabular}}
\end{table}

{\small
\bibliographystyle{ieee_fullname}
\bibliography{report}
}

\end{document}